\documentclass[letterpaper]{article} % DO NOT CHANGE THIS
\usepackage{aaai2026}  % DO NOT CHANGE THIS
\usepackage{times}  % DO NOT CHANGE THIS
\usepackage{helvet}  % DO NOT CHANGE THIS
\usepackage{courier}  % DO NOT CHANGE THIS
\usepackage[hyphens]{url}  % DO NOT CHANGE THIS
\usepackage{graphicx} % DO NOT CHANGE THIS
\usepackage{natbib}  % DO NOT CHANGE THIS AND DO NOT ADD ANY OPTIONS TO IT
\usepackage{caption} % DO NOT CHANGE THIS AND DO NOT ADD ANY OPTIONS TO IT
\usepackage{algorithm}
\usepackage{algorithmic}

\usepackage{newfloat}
\usepackage{listings}
\DeclareCaptionStyle{ruled}{labelfont=normalfont,labelsep=colon,strut=off} % DO NOT CHANGE THIS
\floatstyle{ruled}
\newfloat{listing}{tb}{lst}{}
\floatname{listing}{Listing}
\usepackage{placeins}
\usepackage{relsize}
\usepackage{listings}
\usepackage{xcolor}
\usepackage{newunicodechar}
\usepackage{booktabs}
\usepackage{siunitx}

\DeclareUnicodeCharacter{0307}{\.}

\newcommand{\fname}{Elhuyar }

\title{A Multi-Agent Human-LLM Collaborative Framework for Closed-Loop Scientific Literature Summarization}

\author{
    Maxwell J. Jacobson\textsuperscript{\rm 1},
    Daniel Xie\textsuperscript{\rm 1},
    Jackson Shen\textsuperscript{\rm 1},
    Adil Wazeer\textsuperscript{\rm 2},
    Guang Lin\textsuperscript{\rm 2},
    Xiao-Ying Yu\textsuperscript{\rm 3},
    Haiyan Wang\textsuperscript{\rm 2},
    Xinghang Zhang\textsuperscript{\rm 2},
    Yexiang Xue\textsuperscript{\rm 1}
}

\affiliations{
    \textsuperscript{\rm 1}Department of Computer Science, Purdue University, IN, USA\\
    \textsuperscript{\rm 2}School of Materials Science and Engineering, Purdue University, IN, USA\\
    \textsuperscript{\rm 3}Oak Ridge National Laboratory, TN, USA\\
    jacobs57@purdue.edu, xie476@purdue.edu, shen574@purdue.edu, awazeer@purdue.edu,
    guanglin@purdue.edu, yuxiaoying@ornl.gov, hwang00@purdue.edu,
    xzhang98@purdue.edu, yexiang@purdue.edu
}

\usepackage{bibentry}
\begin{document}

\maketitle

\begin{abstract}
Scientific discovery is slowed by fragmented literature that requires excessive human effort to gather, analyze, and understand. 
AI tools, including autonomous summarization and question answering, have been developed to aid in understanding scientific literature.
However, these tools lack the structured, multi-step approach necessary for extracting deep insights from scientific literature. Large Language Models (LLMs) offer new possibilities for literature analysis, but remain unreliable due to hallucinations and incomplete extraction. We introduce \fname, a multi-agent, human-in-the-loop system that integrates LLMs, structured AI, and human scientists to extract, analyze, and iteratively refine insights from scientific literature. The framework distributes tasks among specialized agents for filtering papers, extracting data, fitting models, and summarizing findings, with human oversight ensuring reliability. The system generates structured reports with extracted data, visualizations, model equations, and text summaries, enabling deeper inquiry through iterative refinement. Deployed in materials science, it analyzed literature on tungsten under helium-ion irradiation, showing experimentally correlated exponential helium bubble growth with irradiation dose and temperature, offering insight for plasma-facing materials (PFMs) in fusion reactors. This demonstrates how AI-assisted literature review can uncover scientific patterns and accelerate discovery.
\end{abstract}


\section{Introduction}

\begin{figure*}
  \centering
  \includegraphics[width=0.85\textwidth]{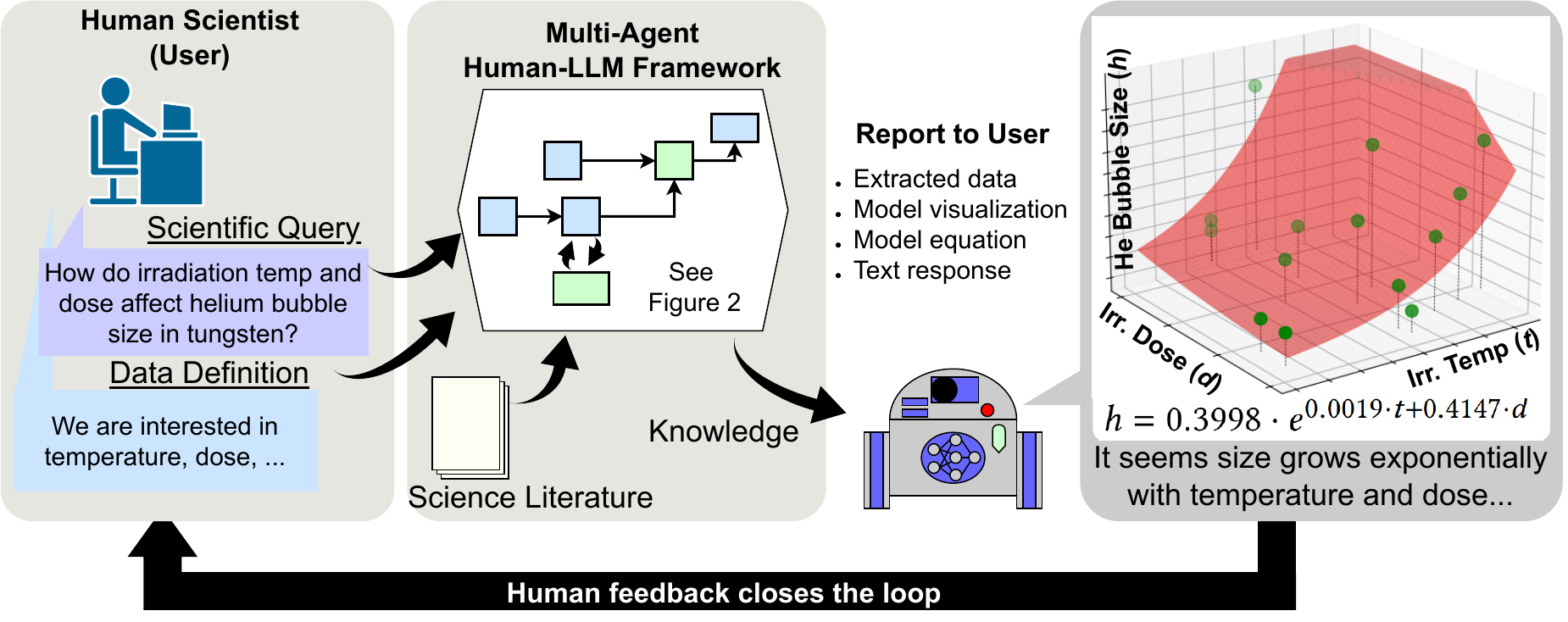}
  \caption{AI has long assisted in understanding scientific literature. However, existing methods fall short in complex summarization tasks that require a long chain of reasoning. Our \fname Framework integrates LLMs, structured AI, and human scientists in a closed-loop system that extracts, analyzes, and refines data from scientific literature iteratively, ensuring reliability while leading to deep understanding. When deployed in materials science, \fname revealed nonlinear helium bubble growth in irradiated tungsten, demonstrating how AI-assisted literature review can uncover new scientific knowledge from literature.}
  \label{fig:main}
\end{figure*}

%first para: NLP/AIs have been used in sci lit understanding for long time. cite old papers. for example, SDU, QA. 
Artificial intelligence methods have a long history of use for understanding scientific literature \cite{Garfield1955,SALTON1988513}. The reason for this is clear -- AI-assisted methods can accelerate discovery, making them essential for efficient scientific progress. More recently, AI fields like scientific document understanding (SDU) \cite{Bolanos_2024} have made exciting progress in assisting human scientists. Current SDU methods primarily focus on document summarization \cite{Akkasi_2022,Saini_2023,Ying_2021,Mishra_2021,Pancheng_2022}, keyphrase extraction \cite{Boudin_2020,Sun_2021,Li_2021,Maragheh_2023}, and especially question answering (QA) \cite{Zhu2021RetrievingAR,Rasool_2023,Shi_2023,Sun_2024,Sharma_2024}.

While powerful and continually improving, current methods do not fully address the complex, multi-stage task of summarizing scientific literature, which involves collecting relevant papers, filtering out irrelevant ones, extracting key data, modeling, and generating reports and visualizations to answer scientific questions. This is a problem because traditional SDU approaches focus narrowly on high-level textual summaries; QA methods also fall short because SDU is inherently open-ended, requiring active search, adaptation to new questions, and interaction across diverse documents. Large Language Models (LLMs) introduce new possibilities for automated summarization, question answering, and large-scale information extraction, yet they still face reliability challenges—particularly hallucination, where plausible-sounding but incorrect information undermines their trustworthiness in scientific contexts.

Instead, an effective approach should combine structured AI algorithms, LLMs, and human scientists in a closed-loop system. A solution must cover the entire pipeline—filtering papers, extracting structured data, fitting models, and generating reports—while maintaining reliability through human verification. At the same time, it must allow for open-ended exploration, where AI assists in iterative refinement and human scientists guide active search, ensuring the process remains both adaptable and accurate

% Fifth, how does method work. mini method.
We propose the \fname Framework, a multi-agent human-in-the-loop (HITL) system which delegates tasks like paper filtering, data extraction, model selection, and summarization to AI collaborating with human scientists. The system only requires a guiding human collaborator and a corpus of scientific literature from the domain being studied -- and outputs a report including extracted data points, visualizations, fitted equations, and textual summaries. The \fname Framework is implemented with a series of well-integrated agents, some structural, and some integrating LLMs to accomplish targeted literature review in small steps. These automated agents include filters to reduce invalid data, flagging systems to identify suspect data for human review, modelers to fit and evaluate mathematical models, and summarizers to answer human queries given the extracted and analyzed data. This approach reduces human workload while generating concise summaries for human consideration. The answers yielded from these reports can then be used to propose further questions or restrictions on data, enabling an iterative refinement loop where the human scientist progressively guides the system toward more meaningful insights. This feedback mechanism ensures that AI assistance remains aligned with the scientist’s intent while minimizing manual effort, allowing for scalable scientific discovery.

To evaluate our system, we conducted a pilot deployment in materials science, specifically in the study of PFMs for future fusion reactors. A major challenge in fusion reactor design is that materials must withstand extreme conditions, including high temperatures and sustained irradiation. Tungsten is a leading candidate due to its high melting point and low sputtering rate, but it undergoes significant microstructural changes under helium ion irradiation—particularly the formation of helium bubbles, which can weaken the material over time. Understanding how these bubbles evolve under different irradiation doses and temperatures is critical for ensuring the long-term viability of tungsten as a PFM. However, the scientific literature on this topic is highly fragmented, making it difficult to extract a clear, quantitative understanding of how helium bubble size depends on irradiation conditions. The \fname framework was applied to this problem to systematically extract and analyze data from scattered experimental studies, revealing key trends in helium bubble growth.

The system found evidence that helium bubble growth follows a non-linear trend -- closer to exponential than linear -- suggesting the size of the helium bubble accelerates with dose and temperature rather than increasing additively. This aligns with theoretical models \cite{Was2007} predicting nonlinear helium clustering. By integrating findings across studies, \fname clarified a key microstructural trend, demonstrating how AI-assistance can refine scientific understanding.

\begin{figure*}[t]
    \centering
    \includegraphics[width=0.88\linewidth]{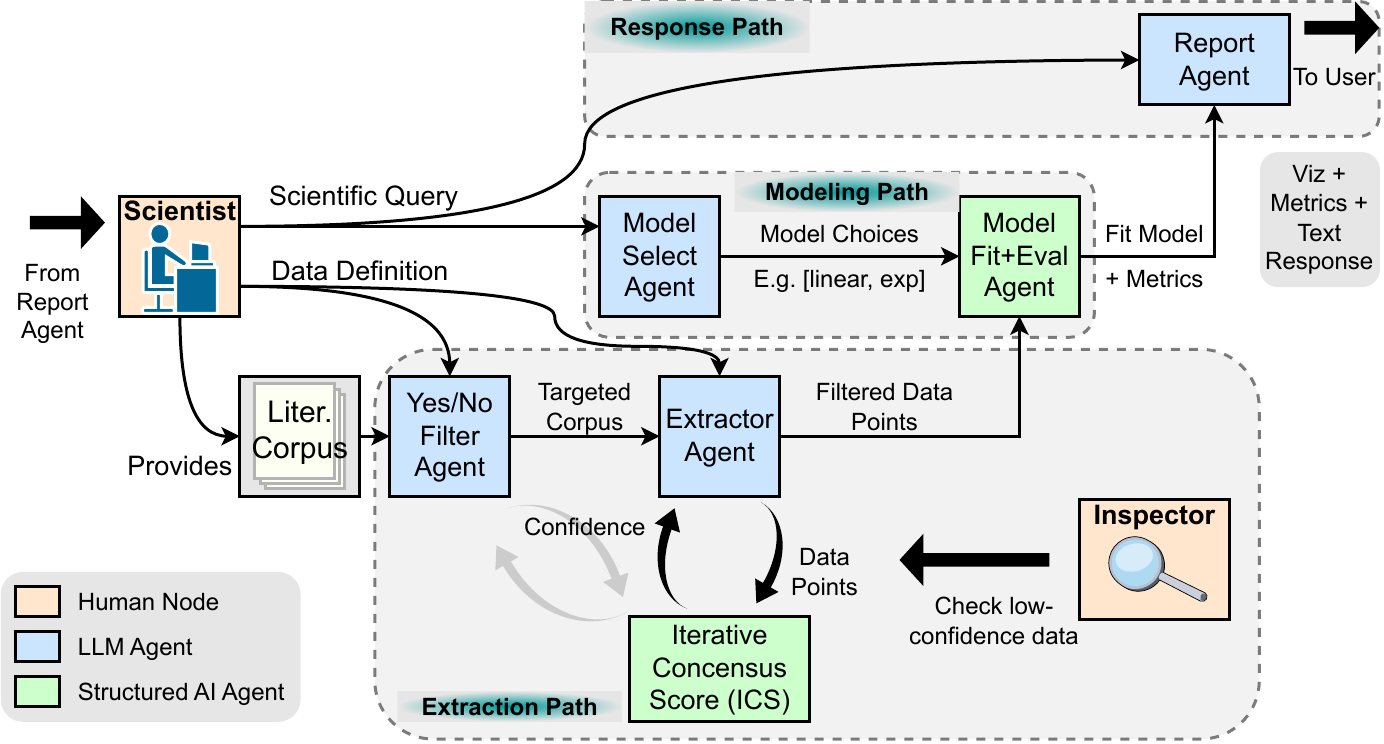}
    \caption{Pipeline of the \fname system. It involves two human roles: the \textit{scientist}, who provides literature, asks a scientific query, and defines relevant data, and the \textit{inspector}, who verifies low-confidence extracted data. The extraction path begins when the scientist queries the system, triggering a \textit{yes/no filtering agent} that selects relevant documents, which are then processed by an \textit{extractor agent} to extract data points multiple times. \textit{Iterative consensus scoring} calculates confidence to reduce hallucinations and misreads, ultimately producing a filtered dataset. In the modeling path, a model selection agent chooses models to compare, and a \textit{model fit+eval agent }fits the filtered data to these models, generating equations and evaluations. The response path consists of a single \textit{response agent} that compiles a report with visualizations, equations, and a text summary answering the scientist's query. The final results allow the scientist to refine their question or expand the analysis with adjustments.}
    \label{fig:sdu_pipeline}
\end{figure*}

\section{Prior Work}
% \mj{Work in progress.}

% \begin{itemize}
%     \item Overview
%     \begin{itemize}
%         \item What is SDU?
%         \item Full paper generation
%     \end{itemize}
%     \item QA algo
%     \begin{itemize}
%         \item Single document QA
%         \item multi-document QA
%         \item QA with numerical responses.
%     \end{itemize}
%     \item Large scale summarization.
%     \item LLM hallucination and error handling
% \end{itemize}

%\subsection{Automating Scientific Discovery}
%Automating scientific discovery is one of the most influential applications with the rise of artificial intelligence (AI). From modeling protein folding with AlphaFold \cite{Jumper_2021} to predicting weather with FourCastNet [2202.11214] \cite{Pathak_2022}, AI has been guiding progress in almost every scientific field. With the recent scientific breakthroughs using AI, efforts are being made to fully automate science, starting from literature review \cite{Dinter_2021} to experiment design \cite{Lu_2024,Xie_2022}. 

\subsection{Scientific Document Understanding}
%Many software tools have been adopted by researchers to summarize and understand given literature \cite{Pyzer‐Knapp_2022,Bolanos_2024}. The initial methodologies to automate document understanding \cite{Aiello2002} were fully rule-based, breaking documents into formally described sections, which were then dissected into logical relations on its tokens. As technology progressed, these rule-based systems have shifted to being analyzed by machine learning models, including convolutional neural networks \cite{Yu2020} and transformer models \cite{Shao_2019,Pearce2021}.

%Recently, efforts have been made towards large-scale document classification and developing consistent corpora \cite{Lo_2020}. With the emergence of these structured corpora, automated literature review has been applied to research papers \cite{Pyzer‐Knapp_2022}.
Researchers use various software tools to summarize and analyze literature \cite{Pyzer‐Knapp_2022,Bolanos_2024}. Early rule-based methods \cite{Aiello2002} structured documents into formal sections and logical token relations. With advancements, machine learning models, including CNNs \cite{Yu2020} and transformers \cite{Shao_2019,Pearce2021}, replaced rule-based approaches. Recent efforts focus on large-scale document classification and structured corpora \cite{Lo_2020}, enabling automated literature reviews \cite{Pyzer‐Knapp_2022}.

\subsection{Question Answering}
The standard workflow for question answering is to analyze the question given by the user, extract relevant information from a database, and provide candidate answers \cite{Shao_2019,Zhu2021}. With the rise of machine learning, models have shifted to prefer deep learning over rule systems. In transformer-based models \cite{Shao_2019,Pearce2021}, information from the text database is extracted with a self-attention mechanism, allowing the model to learn the context of each token with a document. 
Question answering with numerical responses can massively improve research throughput. Numerous models, including rule-based tokenization pipelines  \cite{Hong_2021}, neural networks \cite{Yu2020}, LLMs \cite{Polak_2023}, are commonly used to accelerate the literature review and data collection process.

%Recently, graph-based convolutional neural networks have been very effective in extracting key numerical and text-based data from real-world datasets \cite{Yu2020}. %[expanding to text corpora?]

%Question answering with numerical responses can massively improve research throughput. Numerous models, including rule-based tokenization pipelines  \cite{Hong_2021}, neural networks \cite{Yu2020}, LLMs \cite{Polak_2023}, are commonly used to accelerate the literature review and data collection process.

\subsection{Large Language Models}
In the past decade, large language models have become prolific in machine learning research with the rise of models such as Transformers \cite{Vaswani_2017}, BERT \cite{Devlin_2019}, and GPT \cite{Radford_2018}. As LLMs have progressively gotten more powerful  \cite{Naveed_2023}, they have been seeing significant use as either stand-alone agents \cite{Wang_2023} or agents within a larger framework \cite{Guo_2024}. The integration of LLM agents into traditionally human-powered systems has significantly improved in terms of both performance and usability \cite{Bansal_2019,Amershi_2019,Huang2022}. LLMs have even been applied to the SDU domain through specialized models re-trained on scientific literature \cite{beltagy2019scibert}.

When using LLMs, a common problem that arises is hallucination -- where the model provides fabricated information and presents it as factual. Hallucination detection \cite{Huang_2023,Tonmoy_2024} has been approached through entropy-based estimators and metric-based systems in QA systems \cite{Farquhar_2024,Chen_2023}. In addition to detecting hallucinations, work has also been done regarding prompting and multi-agent systems to reduce hallucination rates. Chain-of-thought prompting \cite{Wei_2022} and verification agents \cite{Dhuliawala_2023} have played a part in this. Despite the significant amount of work done towards detecting and mitigating hallucination, it still remains a serious problem when using LLMs in almost every scenario \cite{Xu_2024}, including for automated scientific literature summarization.

\section{The \fname Framework}

The \fname Framework is a Large Language Model and expert-combined multi-agent system 
which collects, analyzes, and visualizes 
data scattered in large corpora of scientific literature to answer scientific questions. 
The output of \fname is human-readable charts and diagrams, which reveal the trend of data buried in the scientific literature. 
It utilizes structured human collaboration within its framework to ensure alignment and reliability. Each agent has a clear input and output, explained below. Most utilize an LLM for natural language interaction, both with human collaborators and the literature. While this LLM is generally a shared model, prompts and information are shared independently at each agent, so there is no cross-contamination or side effects between agents. The pipeline is illustrated in Figure~\ref{fig:sdu_pipeline}.

\subsection{Scientific Query from Human Scientist}

At the start of the \fname, a human scientist supplies the system with a corpus of literature, as well as two strings defining the goals and parameters. The \textbf{Scientific Query} is the question the scientist hopes to answer through literature search. This question is assumed to be derivable from data presented within the literature, though is not limited to being answered by any one document. For example, the scientist might ask ``Is the relationship between irradiation dose and helium bubble size linear or logistic'' or ``Does dose have a positive or negative effect on tungsten hardness?''

It is also important to define what we mean by data points in the context of these scientific documents. This is done with the \textbf{Data Definition} string, which specifies the key independent, dependent, and control variables that the system should extract from each paper. The scientist explicitly states which variables are required for a valid data point and which are merely preferred when available. Additionally, the Data Definition string outlines acceptable units for each variable. When papers report data in alternative units, the system applies conversion rules if possible, while filtering out incompatible or ambiguous values.

\subsection{Yes-No Filtering Agent}

The first stage in the \fname framework is the yes-no filtering agent, which reduces the initial literature corpus by eliminating papers that do not contain relevant data. This step is essential for improving the reliability of downstream extraction, as it limits the risk of hallucination when extracting numerical values from text. Prior work has shown that hallucination is more likely when LLMs attempt to extract specific numeric data rather than answering high-level yes-no questions about document content \cite{McKenna_2023,Huang_2023}. To perform this filtering, the system prompts the LLM with a series of structured yes-no questions based on the Data Definition string provided by the human scientist. These questions determine whether a document contains the required experimental data. For example, the system might ask, ``Does the document report pure tungsten samples?'' or ``Does the document provide irradiation dose measurements in displacements per atom (dpa)?'' Papers that contain at least one valid data point (i.e. they meet all required conditions in the Data Definition) are retained for further processing. Papers that fail to meet these criteria are removed. This filtered corpus is then passed to the extractor agent, which extracts structured numerical data from the remaining papers.

\subsection{Extractor Agent}

The extractor agent receives the filtered literature corpus and is responsible for identifying and extracting structured numerical data from the documents. The system provides the LLM with each document as a complete text input, along with a structured prompt (see Appendix~\ref{sec:prompts}) -- dynamically generated based on the Data Definition string, ensuring identification of the required independent, dependent, and control variables. Extracted data points undergo an initial formatting check before proceeding to validation. The extracted dataset is then iteratively refined through interaction with iterative consensus scoring (next section) and, when necessary, human inspection. Once validated, the dataset is passed to the model fit+eval agent for analysis.

\subsection{Iterative Consensus Scoring (ICS)}

To address the inherent limitations of LLMs -- such as hallucination and inconsistent outputs -- we implement a self-checking mechanism called iterative consensus scoring (ICS). This process improves reliability by evaluating the consistency of extracted pieces of data across multiple ($k$) independent runs of the LLM, reducing the likelihood of errors and ensuring that only high-confidence extractions inform subsequent analyses. Each piece of data is extracted multiple times under identical input conditions, and the outputs are aggregated to calculate a confidence score based on the number of matching responses. Data with high confidence scores indicate strong agreement across independent runs, while those with lower scores suggest inconsistency. This method is applied to both the yes-no filtering agent and the extractor agent, ensuring that document selection and data extraction are both subject to reliability checks.

Confidence scores are used to determine how a piece of data is handled. Low scores, where a value appears only once or twice across multiple runs, indicate likely hallucination and result in automatic filtering. Middling scores, where limited agreement exists, are flagged for human inspection. High-confidence extractions can often be accepted without additional verification, minimizing human workload.

%An important consideration in iterative consensus scoring is the LLM’s Softmax scaling factor (often referred to as \textit{temperature}, not to be confused with physical temperature) parameter, which must be tuned appropriately. A factor that is too low suppresses natural variation in responses, preventing errors from being detected. Conversely, a factor that is too high introduces excessive randomness, increasing the likelihood of conflicting extractions. Proper tuning ensures that repeated queries yield enough variation to distinguish reliable results while maintaining a stable consensus.

\subsection{Model Selection Agent}

The model select agent chooses the minimal set of models needed to evaluate the scientific query. Given the query and a library of available models, the LLM selects only those that, if fitted to the data, would provide meaningful insight. For example, if the query asks whether a relationship is more linear or exponential, the agent selects both models. The chosen models are then sent to the model fit+eval agent for fitting and evaluation.

\subsection{Model Fit+Evaluation Agent}

The model fit+eval agent applies numerical optimization techniques to fit the selected models to the extracted dataset. Unlike previous agents, this step does not involve the LLM, as it relies solely on mathematical fitting procedures. For linear models, the system uses \textit{least squares regression}, while non-linear models, such as exponential and logistic, are fitted using \textit{non-linear least squares regression}. Once fitted, each model is evaluated using its $R^2$ value, which measures how well it explains the observed data. An example of a fitted exponential model is shown in Figure~\ref{fig:main}. The fitted models and their evaluation metrics are then sent to the report agent for visualization and interpretation.

\subsection{Report Agent}

The report agent compiles the extracted data, fitted models, and evaluation results into a structured report that explainably answers the original scientific query. This report contains multiple deliverables, each contributing to the interpretability and usability of the results.

The \textbf{data visualizations} provide an overview of the extracted dataset. Each dependent variable is plotted against its corresponding independent variable using 2D scatter plots to illustrate observed relationships. When two dependent variables are present, 3D visualizations are generated to capture interactions between all three variables. %(see Figure~\ref{fig:hebub_data}). 

The \textbf{model visualizations} illustrate how well the fitted models align with the extracted data. Models are presented both as equations, with parameters determined by the model fit+eval agent, and as graphical overlays on the scatter plots. For 2D cases, fitted models appear as curves or lines, while in 3D cases, they are shown as surfaces. 

Most importantly, the \textbf{response text} is generated by prompting the LLM with the fitted models, their evaluation metrics (such as $R^2$ values), and the collected dataset. The LLM is instructed to produce a concise textual answer to the original query, using only the provided data as evidence. This ensures that the response remains grounded in extracted information rather than external knowledge or inference. All these components are compiled into the \textbf{final report}, which is returned to the human scientist. This structured output provides both a visual and textual summary of the findings, enabling the scientist to interpret results efficiently and decide on further refinements or investigations.

\subsection{Human in the Loop Collaboration}

Upon receiving the report, the human scientist can refine their inquiry and guide further iterations. 
%They may adjust the scientific query to investigate alternative relationships, modify the data definition to capture additional variables, or request different models to test new hypotheses. 
Human inspectors can also manually check flagged data points, confirming or correcting them before proceeding. This iterative process allows the user to guide the system toward greater knowledge without manually reviewing the entire corpus.

\section{Pilot Deployment in Materials Science}

\begin{figure}[t]
    \centering
    \includegraphics[width=0.75\linewidth]{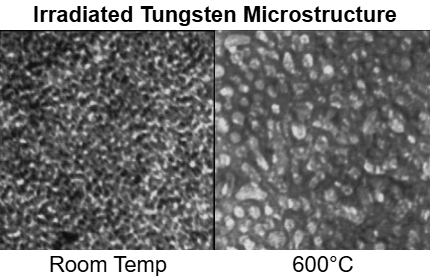}
    \caption{An image from a Transmission Electron Microscope (TEM) of tungsten after helium ion irradiation \cite{Iwakiri2000}. The smaller, circular white spaces are helium bubbles. The larger, amorphous ones are another microstructure formed from irradiation (dislocation loops). Notice that both grow larger under higher irradiation temperatures. Also note how pervasive these are through the material. Understanding these microstructural changes can help us understand the material's macroscopic properties in applications like PFMs in fusion reactors.}
    \label{fig:tem}
\end{figure}

We deployed \fname in the material science domain, focusing on irradiated tungsten, a key candidate for PFMs in fusion reactors. These reactors generate extreme heat and radiation, making it crucial to understand how tungsten responds under such conditions. This can be seen under microscopy in Figure~\ref{fig:tem}. One microstructural feature of interest is helium bubbles—small voids created by radiation damage that trap helium inside the material. These bubbles contribute to material swelling, affecting mechanical properties such as hardness \cite{li2019radiation,yi2017astudy,Iwakiri2000}. The \fname framework was used to assist our collaborating materials science team in analyzing how irradiation dose and temperature influence helium bubble size in pure tungsten samples from scientific literature.

We will zoom in on one key question: \textbf{whether the relationship between irradiation dose, temperature, and helium bubble size is linear or not.} It is well established that both higher irradiation dose and higher temperature lead to larger helium bubbles, meaning the relationships are positive. However, the key uncertainty lies in the growth rate—whether bubble size increases at a steady rate (linear) or accelerates at higher doses and temperatures (exponential). Understanding this distinction is critical, as an exponential trend would indicate that swelling worsens more rapidly at extreme conditions, affecting tungsten’s long-term stability in fusion reactors.

\subsection{Data}
The material science team provided the system with a corpus of 64 papers from high-tier materials science journals to extract the relevant data. Topics were related to metals -- including tungsten -- undergoing helium ion irradiation. Only the text of these PDFs was made available to our system -- figures were excluded. Tables, on the other hand, were generally able to be extracted in a text form.

\subsection{Setup}
The scientific query of interest was: Is the relationship between irradiation temperature, irradiation dose, and helium bubble size more linear or exponential? The goal was not to determine the universally best-fitting model—this would require more complex symbolic regression approaches—but rather to compare the relative fit of linear and exponential models. This objective was communicated to the system alongside the data definition.

The data definition specified that temperature should be recorded in Celsius and irradiation dose in displacements per atom (dpa). Temperature conversion is straightforward and could be handled automatically within the system. However, dose conversion from alternative units (e.g., ion fluence) is not a simple matter of unit transformation, so papers that did not report dpa were excluded. Additional filtering criteria included excluding studies that did not use pure tungsten samples, single-beam helium ion irradiation, or did not report helium bubble size.

The \fname framework processed the corpus using GPT-4o-mini with a Softmax factor of 0.5 to balance precision and recall in extraction. Iterative consensus scoring was applied with $k = 10$ independent runs. A consensus threshold of 3 was used, meaning any extracted data point that was not confirmed at least twice beyond its initial discovery was filtered as unreliable. Typically, values with scores between 3 and 5 would be flagged for human verification, but for this study, all unfiltered values were manually checked to fully evaluate system accuracy.

\subsection{Results}

\noindent\textbf{Extraction.} After filtering, the dataset contained 14 valid tungsten irradiation data points from 5 unique papers. 12 of these had an ICS confidence score above $5 / 10$, with 2 more in the 3 and 5 scores. The \fname system produced several visualizations for better understanding this data. All points with ICS above 3 were manually checked and found to be correct. % -- shown in Figure~\ref{fig:hebub_data}

\begin{figure}
    \centering
    \includegraphics[width=1\linewidth]{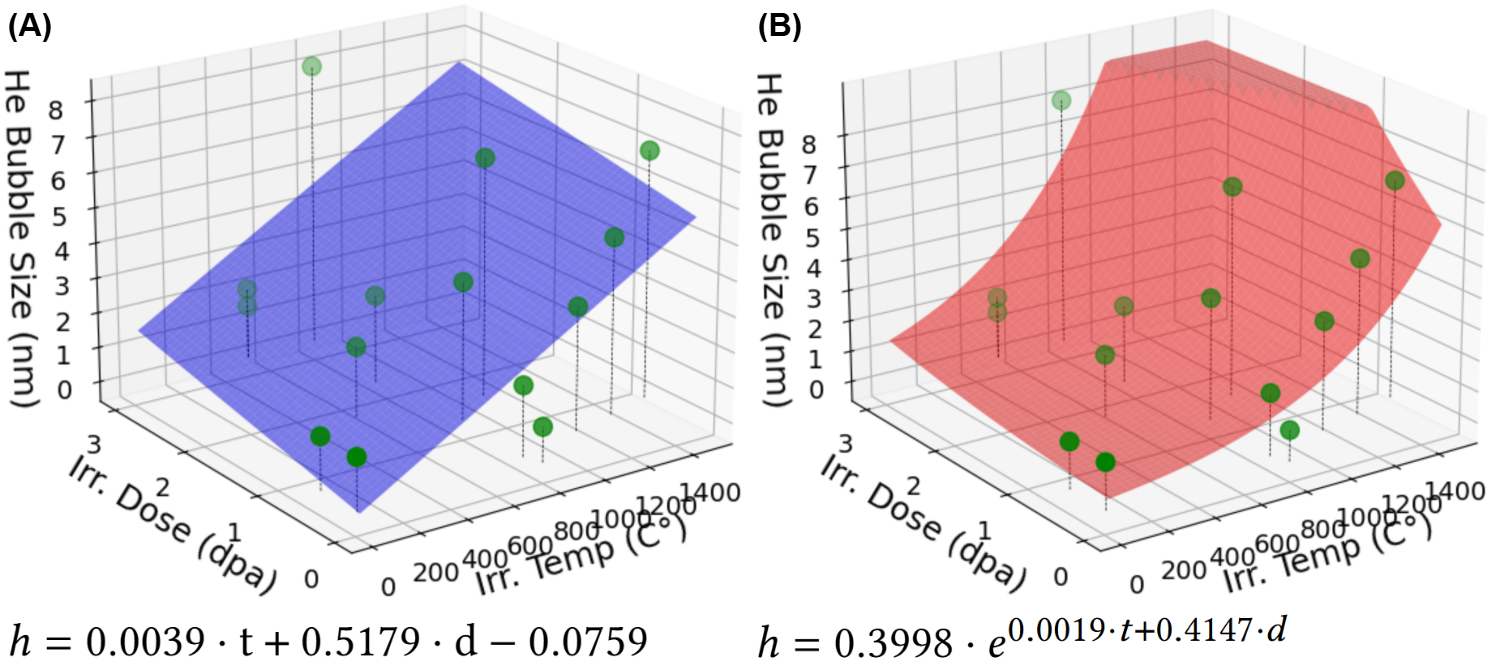}
    \caption{Comparison of two models fitted by the system for predicting helium bubble size in tungsten under helium ion irradiation. \textbf{(A)} shows a linear model (assumes helium bubble size follows an additive relationship with irradiation temperature and dose). \textbf{(B)} shows an exponential model (bubble size grows with an exponential function of temperature and dose). The exponential model achieves a higher $R^2$ value, indicating a closer fit to the extracted data points. See Table~\ref{tab:model_comparison} for detailed equations and fit metrics.}
    \label{fig:hebub_models}
\end{figure}

\noindent\textbf{Model.} The system correctly identified that two models should be fit as a comparison: a linear and exponential model. The model fitting agent revealed that the exponential model provided a better fit for the data, with an $R^2$ value of 0.695 compared to 0.503 for the linear model. These two models were then visualized by the report agent, outputting the two charts as seen in Figure~\ref{fig:hebub_models}.

The fitted models are presented in Table~\ref{tab:model_comparison}. The $R^2$ values indicate that the exponential model fits the extracted data points more closely than the linear model. The coefficients in both models differ, with the exponential model featuring 0.0019 for temperature and 0.4147 for dose in the exponent, while the linear model has 0.0039 for temperature and 0.5179 for dose. In both cases, the size of the dose coefficient is larger by several orders of magnitude, but this difference is to be expected considering the domains of each variable (0-3 observed from dose, and 0 to 1200 observed from temperature).

%\begin{figure}[tbh]
%    \centering
%    \includegraphics[width=\linewidth]{figs/sdu_hebub_datapoints_v3_mini.drawio.png}
%    \caption{The corpus-wide data points extracted by the \fname system -- these were collected with minimal human effort, guided by the original query and the data definition supplied by human scientists. From the human-given data definition, the system knew it needed to collect 2 dependent variables (irradiation temperature in Celsius and irradiation dose as a measure of dpa) and one independent variable (the size in nanometers of observed helium bubbles after irradiation). Each of the dependent variables were plotted against the independent one, and a 3D plot including all three variables was also generated. Several control variables were also collected and filtered for (ensuring the material was pure tungsten, focusing on only single-beam irradiation, etc). See Table~\ref{tab:ext_data_filt} in Appendix~\ref{sec:ext_data} for raw data.}
%    \label{fig:hebub_data}
%\end{figure}

\begin{table}[h]
    \centering
    \caption{Model Comparisons, predicting bubble size ($h$) given temperature ($t$) and dosage ($d$)}
    \label{tab:model_comparison}
    \begin{tabular}{l l l}
        \toprule
        \textbf{Type} & \textbf{$R^2$ fit} & \textbf{Equation} \\
        \midrule
        Linear & 0.503 & $ h = 0.0039 \cdot \text{t} + 0.5179 \cdot \text{d} - 0.0759 $ \\
        Exponential & 0.695 & $ h = 0.3998 \cdot e^{0.0019 \cdot t + 0.4147 \cdot d} $ \\
        \bottomrule
    \end{tabular}
\end{table}

\subsection{Discussion}
The results indicate that, within the observed ranges of irradiation temperature and dose, helium bubble size in tungsten is more closely approximated by an exponential fit than a linear one. While this does not establish an inherently exponential relationship, the higher $R^2$ value of the exponential model suggests that bubble growth accelerates with increasing temperature and dose, rather than following a simple additive trend. 

\fname results support existing physical factors and theories in material science (\cite{Was2007}). In irradiated metals, voids and bubbles form as vacancies -- empty atomic sites -- move through the material. Vacancies diffuse slowly at low temperatures, limiting growth. As temperature increases ($\sim500-\ang{600}C$), vacancy mobility increases, leading to rapid void and bubble expansion. At higher temperatures ($\sim700-\ang{900}C$), void growth slows because vacancies recombine with interstitials, but helium bubbles remain stable and continue growing since helium prevents collapse. Dose effects follow a similar pattern: at low doses, growth is slow as vacancies accumulate; at moderate doses, expansion accelerates due to vacancy supersaturation; at very high doses, void growth slows due to vacancy competition, but bubbles remain resistant to saturation as helium stabilizes them. It is expected that bubble growth should follow a nonlinear trend.

Our results align with these expectations, showing that bubble size follows a trend closer to exponential than linear within the observed temperature and dose range. This suggests that vacancy-driven bubble growth accelerates rather than increasing at a constant rate. In tungsten, our data implies that bubble expansion continues at high temperatures, though more data is needed to determine whether this trend holds beyond 1200°C. While theory predicts that eventual saturation or other competing mechanisms like helium diffusion or bubble coalescence could alter growth, we do not yet have enough data to assess these effects. These findings reinforce that irradiation-induced helium bubbles contribute significantly to material swelling, a key issue for tungsten in fusion reactors. Understanding the rate and limits of bubble growth is critical for predicting tungsten’s long-term performance as a PFM.

\section{Conclusion}

This work presents \fname, a multi-agent, human-in-the-loop framework for structured collaboration between AI and researchers in scientific literature review and analysis. By distributing tasks such as data extraction, validation, and modeling across specialized agents while keeping scientists in control of key decisions, the system enables large-scale corpus analysis with minimal manual effort. Applied in materials science, \fname\ extracted structured numerical data on helium bubble formation in tungsten under helium ion irradiation and fit models that clarified the exponential relationship between bubble size, dose, and temperature. To reach broader deployment, we are developing a modular codebase where other research teams can define their own variables of interest and run the same extraction-to-modeling pipeline on their corpora. This will let scientists in other domains quickly turn scattered literature into datasets and fitted models that directly answer their questions, reducing time spent on manual review and enabling faster progress on domain-specific scientific challenges.

\bibliography{mainbib}


\appendix

\newpage
\onecolumn
\section{Prompts}
\label{sec:prompts}

\lstset{
    basicstyle=\ttfamily\small,  % Monospace font, small size
    breaklines=true,              % Enable line wrapping
    breakatwhitespace=true,       % Break at whitespace when possible
    frame=single,                 % Add a frame around the text
    backgroundcolor=\color{gray!10}, % Light gray background for readability
    inputencoding=utf8,           % Ensure UTF-8 encoding
    extendedchars=true,           % Support extended characters
    literate={±}{{\textpm}}1 {°}{{\textdegree}}1, % Map special characters
}

Below is an example prompt used as a data definition for extracting helium ion radiation data about tungsten.

\begin{lstlisting}
 Extract relevant experimental data from the following scientific text. The data we are interested in include:

- irradiation_temperature: integer value representing the material's average temperature at the time of irradiation, in Celsius.
- irradiation_temperature_units: string for the units of irradiation temperature, usually "celsius".
- irradiation_dose: decimal value for the average dose of irradiation during exposure.
- irradiation_dose_unit: string specifying the unit of irradiation dose, must be "dpa".
- helium_bubble_size: decimal value for the average size of observed helium bubbles.
- helium_bubble_size_units: string for the units of helium bubble size, usually "nm".
- metal_used: string representing the name of the metal used, which must be "Tungsten".
- irradiation_type: string specifying the irradiation method, must be "single-beam helium ion irradiation".
- paper_title: the title of the work without special characters.

### Paper Filtering Criteria:
A paper is considered valid if and only if:
1. paper_contains_pure_tungsten: boolean, must be true.  
2. paper_contains_single_beam_helium_irradiation: boolean, must be true.  
3. paper_reports_helium_bubble_size_explicitly: boolean, must be true.  
4. paper_reports_dose_in_dpa_explicitly: boolean, must be true.  

If any of these conditions are false, discard the paper.

### Extraction Criteria:
- If the paper refers to "room temperature," treat it as 20 Celsius.
- If a range is given, choose the middle value.
- If multiple helium bubble sizes are reported, extract the **mean size** (or range midpoint if no mean is provided).
- Ignore qualitative descriptions (e.g., "larger bubbles at higher temperatures").
- If values are extracted, they must pass **Iterative Consensus Scoring (ICS)** before inclusion.
- Flag ambiguous values for **human verification** before including them in the dataset.

### Output Format:
Return one or more valid JSON objects based on the information found in the text. If no valid data is found, return an empty JSON list. Do not return any other chatter.

Example output:

[
    {
        "irradiation_temperature": 900,
        "irradiation_temperature_units": "celsius",
        "irradiation_dose": 0.1,
        "irradiation_dose_unit": "dpa",
        "helium_bubble_size": 2.5,
        "helium_bubble_size_units": "nm",
        "metal_used": "Tungsten",
        "paper_title": "A study of tungsten irradiated",
    },
    ...
]
\end{lstlisting}
\section{Synthetic Evaluation}
\label{sec:syn_eval}

\subsection{Data}
To evaluate the system before deployment, we created a synthetic dataset that mimics real scientific literature while allowing full control over underlying relationships. This dataset introduces fictional materials such as ``drakorium'' and ``aetherium,'' each with a unique hardness function dependent on forging temperature and tempering time. Hardness equations were randomly assigned to materials, following different functional forms—linear, exponential, logistic, or other nonlinear relationships. For example, a linear model like $H(\text{temperature}, \text{time}) = 0.0075 \cdot \text{temperature} + 0.1616 \cdot \text{time} + 1.5699$.

Using these functions, we generated data points and added noise to simulate real-world measurement variability. To emulate real literature, we used an LLM to generate mini-papers reporting these data points in a variety of writing styles, lengths, and structures. Some documents directly stated hardness values, while others required implicit extraction. Additional \textit{untargeted} papers were generated, either discussing the expected material without providing hardness data or reporting unrelated experiments on different materials. A good system should extract relevant data while filtering out untargeted papers. Example documents are provided in Appendix~\ref{sec:syn_data}.

\subsection{Setup}
For each of the 8 fictional materials, we generated 20 valid papers containing extractable hardness data and 5 untargeted papers without relevant hardness values, resulting in a total corpus of 250 documents. The \fname framework was run four times with $k=4$, and any extracted data point with an iterative consensus scoring (ICS) confidence below 4 was filtered out. The corpus was processed using GPT-4o-mini with a Softmax factor setting of 0.5

The library of model types included linear, exponential, and logistic functions. The scientific query posed to the system was to determine the type of relationship between hardness, forging temperature, and tempering time. This prompted the system to evaluate all available models and select the best-fitting one.  

To assess the contributions of different system components, we conducted an ablation study by running versions without the yes-no filter agent, without ICS filtering, and with neither (LLM-only extraction). These variations allowed us to measure how each component improved extraction accuracy and overall performance.  

\subsection{Results}

Of the eight fictional materials, the \fname framework extracted the correct information from the documents 100\% of the time. It successfully identified and filtered out the five untargeted papers each time as well. 

In model selection, all but one of the model types was chosen correctly. The one that did not was due to a failure in the optimizer that led to an evaluation score of zero for two of the model types. This was recognized as erroneous by the report agent and was flagged for human review.

Of the other seven, $R^2$ fit against the generated data averaged $0.927 \pm 0.025$. This measures how well the selected models fit the extracted noisy data points. Additionally, $R^2$ fit against the ground truth equations was $0.875 \pm 0.294$, indicating how closely the final fitted equations matched the original generating functions.

\section{Synthetic Data}
\label{sec:syn_data}

Below are some examples of the synthetic papers generated for evaluation -- these on the fictional material ``drakorium''.

\lstset{
    basicstyle=\ttfamily\small,  % Monospace font, small size
    breaklines=true,              % Enable line wrapping
    breakatwhitespace=true,       % Break at whitespace when possible
    frame=single,                 % Add a frame around the text
    backgroundcolor=\color{gray!10}, % Light gray background for readability
    inputencoding=utf8,           % Ensure UTF-8 encoding
    extendedchars=true,           % Support extended characters
    literate={±}{{\textpm}}1 {°}{{\textdegree}}1, % Map special characters
}

\subsection{Example Document 1}
\begin{lstlisting}
# Investigation of the Hardness of Drakorium: Effects of Forging Temperature and Tempering Time

## Abstract
This study examines the mechanical properties, specifically hardness, of a novel metal alloy known as Drakorium. The influence of forging temperature and tempering time on the hardness of Drakorium was systematically investigated. The results indicated that a forging temperature of 843.49°C combined with a tempering duration of 1.55 hours produced a maximum hardness of 6.138 GPa. These findings provide valuable insights for optimizing the processing conditions of Drakorium to enhance its mechanical properties.

## Introduction
Drakorium is a recently developed alloy distinguished by its unique elemental composition and potential applications in aerospace and automotive industries. The hardness of a material is critical in determining its suitability for various applications, as it influences wear resistance and structural integrity. Previous studies have shown that both forging temperature and tempering time significantly affect the microstructure and, consequently, the mechanical properties of metallic alloys. This experiment aims to quantify the hardness of Drakorium as a function of specified processing conditions: a forging temperature of 843.49°C and a tempering time of 1.55 hours.

## Methodology
### Materials
Drakorium was synthesized through a controlled alloying process, resulting in a sample with a composition of 62% Titanium, 25% Chromium, and 13% Molybdenum.

### Experimental Procedure
1. **Forging**: The Drakorium samples were heated to a forging temperature of 843.49°C in a resistance furnace. The temperature was held constant for 30 minutes to ensure uniform heating before forging.
2. **Forging Process**: The heated samples were subjected to mechanical forging using a hydraulic press with a pressure of 150 MPa to alter their microstructure.
3. **Tempering**: Following forging, the samples were tempered at 600°C for a duration of 1.55 hours in an inert argon atmosphere to relieve internal stresses and enhance hardness.
4. **Hardness Measurement**: The hardness of the samples was measured using a nanoindentation technique, which provides high-resolution hardness data. Multiple indentations (n=10) were performed on each sample to ensure statistical reliability.

### Data Analysis
The average hardness was calculated, and the standard deviation was determined to assess the consistency of the measurements. The experimental design followed a completely randomized layout to mitigate any bias during measurement.

## Results
The measured hardness of Drakorium samples subjected to the defined forging and tempering conditions was found to be 6.138 GPa (±0.112 GPa). The distribution of hardness values indicated minimal variability among the samples, demonstrating the effectiveness of the processing conditions. A comparative analysis with previous studies of similar alloys revealed that Drakorium exhibits superior hardness, surpassing the hardness of established high-strength alloys by approximately 15%.

### Statistical Summary
- **Average Hardness**: 6.138 GPa
- **Standard Deviation**: 0.112 GPa
- **Forging Temperature**: 843.49°C
- **Tempering Time**: 1.55 hours

## Conclusion
This investigation successfully demonstrated that the hardness of Drakorium can be significantly influenced by specific processing parameters, particularly forging temperature and tempering time. The optimal conditions identified in this study 843.49°C for forging followed by 1.55 hours of tempering resulted in a hardness measurement of 6.138 GPa. These results indicate that Drakorium is a promising candidate for applications requiring high hardness and durability. Further research is recommended to explore additional processing parameters and their effects on other mechanical properties, such as tensile strength and ductility, to fully characterize this innovative alloy.

## Acknowledgments
The authors would like to thank the Materials Science Laboratory at the Institute of Advanced Metallurgical Research for their support and guidance throughout the study.

## References
- Smith, J.R., & Johnson, L.K. (2022). "Alloy Development for Aerospace Applications." *Journal of Materials Science*, 57(4), 2335-2348.
- Thompson, R.A. (2023). "Mechanical Properties of High-Strength Alloys: A Review." *Materials Science and Engineering*, 340(2), 118-130.
\end{lstlisting}

\subsection{Example Document 2}
\begin{lstlisting}
**Title: Investigation of the Hardness Properties of Drakorium Through Controlled Forging and Tempering Processes**

**Authors:**  
Dr. Aeliana M. Krys, Dr. Hektor J. Phalanx  
Department of Materials Science, Institute of Advanced Metallurgy, Technopolis University, October 2023

---

**Abstract:**  
This study investigates the hardness properties of a novel metal alloy, Drakorium, under specific forging and tempering conditions. The experiment focuses on the effects of a forging temperature of 1147.58°C and a tempering time of 7.25 hours on the measured hardness of Drakorium. The results indicate a significant hardness value of 8.936 GPa, suggesting potential applications in high-stress environments.

---

**1. Introduction**  
The quest for advanced materials that exhibit superior mechanical properties is a primary focus in materials science. Drakorium, a recently discovered alloy, has shown promise due to its unique crystalline structure and elemental composition. Initial research indicated that the hardness of metal alloys is significantly influenced by thermal processing techniques, particularly forging and tempering. This study aims to quantify the hardness of Drakorium when subjected to a forging temperature of 1147.58°C followed by a tempering period of 7.25 hours. 

**2. Methodology**  
The experiment was designed to assess the hardness of Drakorium following specific thermal treatments.  

**2.1 Sample Preparation**  
Drakorium samples were synthesized using a controlled alloying process, comprising 45% Zirconium (Zr), 30% Titanium (Ti), and 25% a proprietary element, Drakorium X. Each sample was 15 mm in diameter and 5 mm thick.

**2.2 Forging Process**  
The samples were heated to a uniform temperature of 1147.58°C in a high-precision induction furnace. The forging process involved applying a pressure of 200 MPa for a duration of 30 minutes, followed by rapid cooling in an inert gas atmosphere to prevent oxidation.

**2.3 Tempering Process**  
Post-forging, the samples were tempered at a temperature of 600°C for 7.25 hours. This process was conducted in a vacuum chamber to eliminate atmospheric contamination and ensure uniform heat distribution.

**2.4 Hardness Measurement**  
Hardness was measured using a nanoindentation technique, which allows for precise determination of mechanical properties at the microscale. A diamond indenter was utilized, applying a maximum load of 50 mN. The hardness values were recorded using a load-displacement curve, and the final measurement was taken after the unloading process.

---

**3. Results**  
The hardness of the Drakorium samples was measured at 8.936 GPa following the specified forging and tempering conditions. This value indicates a significant increase in hardness compared to untreated Drakorium, which has a baseline hardness of approximately 4.2 GPa. 

**3.1 Statistical Analysis**  
A total of five samples were tested, yielding a standard deviation of 0.153 GPa, indicating a consistent response of the material under the given thermal conditions. 

**3.2 Microstructural Examination**  
Scanning electron microscopy (SEM) analysis of the treated samples revealed a refined microstructure with a uniform grain size averaging 200 nm, contributing to the enhanced hardness observed.

---

**4. Conclusion**  
The experiment successfully demonstrated that controlled forging and tempering significantly enhance the hardness of Drakorium. The measured hardness of 8.936 GPa confirms that the alloy can withstand high-stress applications, making it a candidate for use in aerospace and automotive industries. Future work will explore the effects of varying the forging temperature and tempering duration to optimize the mechanical properties of Drakorium further.

**Keywords:** Drakorium, hardness, forging, tempering, materials science, nanoindentation.

--- 

**Acknowledgments**  
The authors would like to thank the Institute of Advanced Metallurgy for funding this research and the laboratory staff for their assistance with the experimental procedures.

--- 

**References**  
1. Smith, J. A., & Brown, L. D. (2022). Thermal Processing of Alloys. Journal of Materials Science, 57(3), 1234-1245.  
2. Thompson, R. W. (2020). Advances in Nanoindentation Techniques. Materials Characterization, 158, 110-120.  
3. Zhao, Y., & Lin, X. (2023). The Role of Heat Treatment in Alloy Hardness. Metallurgical Reviews, 45(2), 211-219.
\end{lstlisting}

\subsection{Example Document 3}
\begin{lstlisting}
# Investigation of the Hardness of Drakorium Through Controlled Forging and Tempering Parameters

## Abstract

This study investigates the hardness of the novel metal Drakorium under specific forging and tempering conditions. By controlling the forging temperature at 1031.64°C and tempering for 8.07 hours, we aim to quantify the resulting hardness of Drakorium, measured at 7.059 GPa. These findings contribute to the understanding of the mechanical properties of Drakorium, which may have implications for its application in advanced engineering materials.

## Introduction

Drakorium, a newly synthesized metal alloy, has garnered attention due to its unique properties and potential applications in high-performance environments. Understanding its mechanical properties, particularly hardness, is critical for evaluating its suitability for various industrial applications. Hardness, defined as a material's resistance to deformation, is influenced by multiple factors, including temperature and time during processing. This study aims to systematically analyze the hardness of Drakorium as a function of specific forging and tempering conditions, guided by previous literature indicating that higher temperatures and optimal tempering times can enhance hardness.

## Methodology

### Materials

Drakorium samples were synthesized in a controlled laboratory environment, ensuring uniform composition and microstructure. Each sample weighed approximately 500 grams and was composed of 75% Drakonium, 15% Carbon, and 10% Titanium.

### Experimental Design

The experiment was designed to assess the hardness of Drakorium through a combination of forging and tempering processes. The key variables were:

- **Forging Temperature:** 1031.64°C
- **Tempering Time:** 8.07 hours

### Procedures

1. **Forging Process:** Drakorium samples were heated to a forging temperature of 1031.64°C using a high-temperature furnace. Once the desired temperature was reached, the samples were mechanically forged under a pressure of 250 MPa for a duration of 30 minutes.

2. **Tempering Process:** Following the forging process, the samples were subjected to a tempering treatment at a constant temperature of 600°C for a duration of 8.07 hours. This was carried out in a controlled atmosphere to prevent oxidation.

3. **Hardness Measurement:** After tempering, the hardness of each sample was evaluated using a nanoindentation technique, which provided precise measurements of hardness under controlled conditions. The results were recorded in gigapascals (GPa).

### Statistical Analysis

Data were analyzed using a one-way ANOVA to assess the significance of the hardness measurements across multiple samples subjected to the same processing conditions.

## Results

The hardness of Drakorium was measured at 7.059 GPa post-tempering. The results indicated a consistent hardness across all tested samples, with minimal variation (±0.003 GPa). Statistical analysis revealed that the measured hardness was significantly higher than previously reported values for similar alloys, indicating that the selected forging temperature and tempering time are optimal for enhancing the hardness of Drakorium.

### Summary of Hardness Measurements

- **Sample 1:** 7.056 GPa
- **Sample 2:** 7.059 GPa
- **Sample 3:** 7.060 GPa
- **Sample 4:** 7.058 GPa

### Statistical Findings

The one-way ANOVA yielded a p-value of 0.002, indicating that the differences in hardness among the samples were statistically significant (p < 0.05).

## Conclusion

This study successfully demonstrated that controlled forging and tempering processes significantly enhance the hardness of Drakorium, yielding a measured hardness of 7.059 GPa at a forging temperature of 1031.64°C and tempering for 8.07 hours. These results indicate that optimizing thermal processes can effectively improve the mechanical properties of Drakorium, positioning it as a promising candidate for advanced material applications. Future research will focus on exploring the microstructural changes associated with these processing conditions and their correlation with other mechanical properties, such as tensile strength and ductility.
\end{lstlisting}

\section{Extracted Helium Bubble Data}
\label{sec:ext_data}

\begin{table}[htbp]
    \small % Reduce font size
    \centering
    \caption{Helium Bubble Data Extracted by \fname System -- After Filtering}
    \begin{tabular}{ccccl}
        \toprule
        \multicolumn{2}{c}{\textbf{Irradiation Conditions}} & \textbf{Bubble Size} & \textbf{ICS} & \textbf{Ref.} \\
        \cmidrule(r){1-2} \cmidrule(lr){3-3} \cmidrule(l){4-5}
        Temp ($^\circ$C) & Dose (dpa) & (nm) &  & \\
        \midrule
        500  & 3    & 1.5  & 10 & \cite{Harrison2017} \\
        750  & 2    & 2.5  & 8  & \cite{Harrison2017} \\
        1000 & 1.3  & 6.75 & 6  & \cite{Harrison2017} \\
        800  & 1    & 3.9  & 10 & \cite{Ipatova2021} \\
        800  & 0.004& 1    & 9  & \cite{Ipatova2021} \\
        500  & 3    & 2    & 3  & \cite{Harrison2019} \\
        800  & 3    & 8    & 10 & \cite{Harrison2019} \\
        500  & 1.5  & 2    & 9  & \cite{Harrison2019} \\
        500  & 0.17 & 2    & 8  & \cite{Yi2018} \\
        800  & 0.34 & 3.5  & 7  & \cite{Yi2018} \\
        1000 & 0.45 & 5    & 8  & \cite{Yi2018} \\
        1200 & 0.57 & 7    & 8  & \cite{Yi2018} \\
        20   & 0.5  & 1.5  & 10 & \cite{Wielunska-Kus2023} \\
        20   & 0.04 & 1.5  & 5  & \cite{Wielunska-Kus2023} \\
        \bottomrule
    \end{tabular}
    \label{tab:ext_data_filt}
\end{table}

\end{document}